\documentclass[conference]{IEEEtran}
\IEEEoverridecommandlockouts

\usepackage[T1]{fontenc}
\usepackage[utf8]{inputenc}
\usepackage{times}             % Times text (matches the IEEE draft); math stays CM
\usepackage{amsmath,amssymb}
\usepackage{booktabs}
\usepackage{array}
\usepackage{enumitem}
\usepackage{microtype}
\usepackage{xcolor}
\usepackage{tikz}
\usetikzlibrary{positioning,arrows.meta,fit,backgrounds,calc}
\usepackage[hidelinks]{hyperref}
\usepackage{cleveref}
\usepackage{cite}

\newcommand{\code}[1]{\texttt{#1}}
\newcommand{\run}[1]{\texttt{#1}}

\definecolor{mred}{HTML}{9E332A}
\definecolor{mteal}{HTML}{2A6367}
\definecolor{mslate}{HTML}{3C5878}
\definecolor{mink}{HTML}{3A3330}
\definecolor{mredbg}{HTML}{F7EAE8}
\definecolor{mtealbg}{HTML}{E4EFF0}
\definecolor{mslatebg}{HTML}{E7ECF3}
\definecolor{mgreybg}{HTML}{F1EEEC}

\tikzset{
  fig/.style     = {font=\sffamily, line width=0.4pt},
  bx/.style      = {rectangle, rounded corners=1.6pt, draw=mink!55, fill=white,
                    align=center, inner sep=3pt, font=\sffamily\scriptsize,
                    text=mink, minimum height=6mm},
  bxg/.style     = {bx, fill=mgreybg},
  bxr/.style     = {bx, draw=mred, fill=mredbg, text=mink},
  bxt/.style     = {bx, draw=mteal, fill=mtealbg, text=mink},
  bxs/.style     = {bx, draw=mslate, fill=mslatebg, text=mink},
  ttl/.style     = {font=\sffamily\scriptsize\bfseries, text=mink},
  ttlr/.style    = {font=\sffamily\scriptsize\bfseries, text=mred},
  ttlt/.style    = {font=\sffamily\scriptsize\bfseries, text=mteal},
  sub/.style     = {font=\sffamily\tiny, text=mink!70, align=center},
  grp/.style     = {rectangle, rounded corners=2.5pt, draw=mink!35, inner sep=5pt},
  grpr/.style    = {grp, draw=mred!70, dash pattern=on 2pt off 1.5pt},
  ar/.style      = {-{Stealth[length=4pt,width=3pt]}, draw=mink!75, line width=0.45pt},
  arr/.style     = {ar, draw=mred},
  art/.style     = {ar, draw=mteal},
  ard/.style     = {ar, draw=mink!45, dash pattern=on 2pt off 1.5pt},
}

\title{Agent Mesh: Reliability Primitives for Non-Idempotent Agent Delegation\\[3pt]
{\large\normalfont Identity Adequacy and Evidence Adequacy}%
  \thanks{Preprint for arXiv (cs.AI; cross-listed cs.SE, cs.DC, cs.MA), August 2026.
  Supplementary material accompanying the preprint documents the platform's evidence
  boundaries, recovery machinery, and incident corpus.}}

\author{%
\IEEEauthorblockN{Mazhar Shaikh\textsuperscript{1} \quad
Anurag Rajkumar Bombarde\textsuperscript{1} \quad
Harshal Pathak\textsuperscript{2}%
\thanks{\textsuperscript{1}Primary authors; contributed equally.
\textsuperscript{2}Contributing author.}}
}

\begin{document}
\maketitle

\begin{abstract}
Autonomous agents are increasingly deployed to perform bounded software tasks---generating a
component, running a suite, repairing a defect---under an orchestrator that retries, resumes, and
budgets them. The reliability machinery such orchestrators reach for is the service mesh's: retry,
timeout, and error-rate circuit breaking. We report a failure study of a production agentic
software-delivery platform (66{,}185 lines, 59 modules) over 147 numbered incidents spanning 81
identified runs, each recorded with a measured cost and, in the majority of cases, a mutation proof
that reverting the fix reproduces the failure. The study finds that all three assumptions those
primitives rest on are violated in practice, and quantifies the consequences: a loop of
\emph{fifty-four consecutive successful} tool calls that no error-rate breaker could see; a progress
signal computed over an identifier that was constant by construction, which guaranteed a false trip
on the third repair round and drove one run from six of six components to three; twenty-one events
accumulated across six invocations of one delegation, making a correct and demonstrably idempotent
component unwinnable; a misrouted failure that woke five components for a two-component fault and
left three bystanders regressing working code; and twelve distinct incidents in which the
enforcement layer blocked \emph{correct} work, the most expensive costing 107 agent turns and zero
accepted writes. We find one cross-cutting cause and its dual. \emph{Identity adequacy}: in five separate
subsystems an identity that failed to discriminate produced a confident wrong answer, and two of
them derived the corrective rule independently. \emph{Evidence adequacy}: a reliability decision
may be taken only on evidence capable of moving, attributable to what it measures, and
deterministic under identical conditions. From the findings we derive seven reliability
primitives whose enforcement unit is the delegation rather than the message, report what changed when
each was deployed, and specify the controlled evaluation the study motivates but does not itself
constitute.
\end{abstract}


\section{Introduction}

An \emph{agent delegation} is the assignment of a bounded software task to an autonomous agent that
interleaves reasoning with tool invocation to accomplish it~\cite{react}, composing its actions at
inference time. We are deliberately agnostic to how that loop is expressed: nothing below depends on
a particular framework, only on a delegation being effectful, generating its operation set at
inference time, costing tokens whether or not its work is kept, and being retried, resumed, or
repaired by a peer. Orchestrators that schedule such delegations at scale need
reliability machinery, and the machinery they inherit is the service mesh's: bounded retry on
failure, wall-clock timeout, and a circuit breaker driven by error rate. Those primitives rest on
three assumptions---that requests are idempotent or can be made so with a developer-supplied key,
that latency signals failure, and that a discarded request costs nothing.

This paper reports what happens when those assumptions meet real agent traffic. Our subject is a
production agentic software-delivery platform that implements, tests, and repairs multi-stack
codebases from a requirements definition, and its recorded failure corpus: 147 numbered incidents
across 81 identified runs. We did not construct the corpus to test a hypothesis; it accumulated as
an operational record, and the analysis is retrospective.

Two incidents introduce the shape of the problem.

An independent-verifier agent issued the same tool call, with one distinct payload,
\textbf{fifty-four times over eleven minutes}, and stopped only when a human killed the run. Every
one of those calls returned success. No error path was ever reached, so no error-rate breaker could
have fired; the step budget, sized from the workload, bought a proportionally large licence to spin.

Separately, a service's event log---kept deliberately outside the transactional workspace so a
cleanliness check would not revert it, and therefore outside everything that cleans---accumulated
\textbf{21 events across four hours and six check invocations}. A test asserting that exactly one
event had been published failed with three, having observed effects committed by \emph{previous
invocations of the same delegation}. The component was unwinnable however correct its code was. The
service's own idempotency was intact: each of the six invocations published exactly three events.
The duplication was in the ledger, not the producer.

\paragraph{Contributions.} (1) A failure study of a production agentic delivery platform, with the
incident corpus, its collection method, and its costs (\Cref{sec:system,sec:method}). (2) Seven
findings, each supported by measured incidents, covering how agents fail, why error rate and wall
clock are the wrong signals, how effects escape transactional containment, how failure attribution
damages correct work, and how the enforcement layer becomes a primary source of outages
(\Cref{sec:findings}). (3) A cross-cutting result---\emph{identity adequacy}---that unifies five
otherwise unrelated subsystem failures, and which two subsystems derived independently
(\Cref{sec:spine}). (4) \emph{Agent Mesh}: the set of reliability primitives the findings imply, defined against an
abstract delegation interface so it is not specific to our architecture, together with what changed
when each was deployed (\Cref{sec:primitives,sec:outcomes}). (5) The controlled evaluation the study
motivates, stated with a designed kill criterion, together with an explicit account of what is not
yet built (\Cref{sec:eval}).

\paragraph{What this paper is not.} It is not a controlled evaluation, and we are careful throughout
about which claims the design supports: findings about \emph{what fails and why} are evidenced by
recorded incidents; claims about \emph{how much the proposed primitives help} are not made. The incidents are observed,
not induced; the platform is one system; and the primitives are reported with deployment outcomes
rather than with a baseline comparison. \Cref{sec:eval} specifies the controlled study, and
\Cref{sec:threats} states what the observational design cannot support.

\section{The system under study}\label{sec:system}

The subject is a deterministic-DAG agentic delivery platform in production use. A run compiles a
reviewable project declaration into a set of delegations, schedules them under a dependency DAG with
failure isolation between components, and drives each through a lifecycle of test design, test
preflight, red validation, implementation, local verification, oracle qualification, and
cross-service acceptance. The delivery agent measured throughout is 66{,}185 lines across 59
modules.

Three properties make it a useful measurement subject. First, delegations are \emph{effectful}: they
write files, seed databases, publish events, install packages, and call model providers. Second, the
set of effectful operations is \emph{not known when the code is written}---the agent generates it at
inference time---so there is no site at which a developer could attach an idempotency key. Third,
delegations are \emph{expensive}: tokens are spent whether or not the work is kept, so a discarded
delegation is a real loss rather than a freed connection.

The agent's tool surface is closed: exactly seven tools---read, list, grep, write, edit,
run-a-named-check, done---and no shell. The only routes to a subprocess are naming a declared check
specification, or writing a file, which implicitly triggers the stack's declared post-write hooks.
This matters for the study because it makes effects observable at the tool boundary mechanically
rather than heuristically, and it is why the observations below are attributable rather than
inferred.

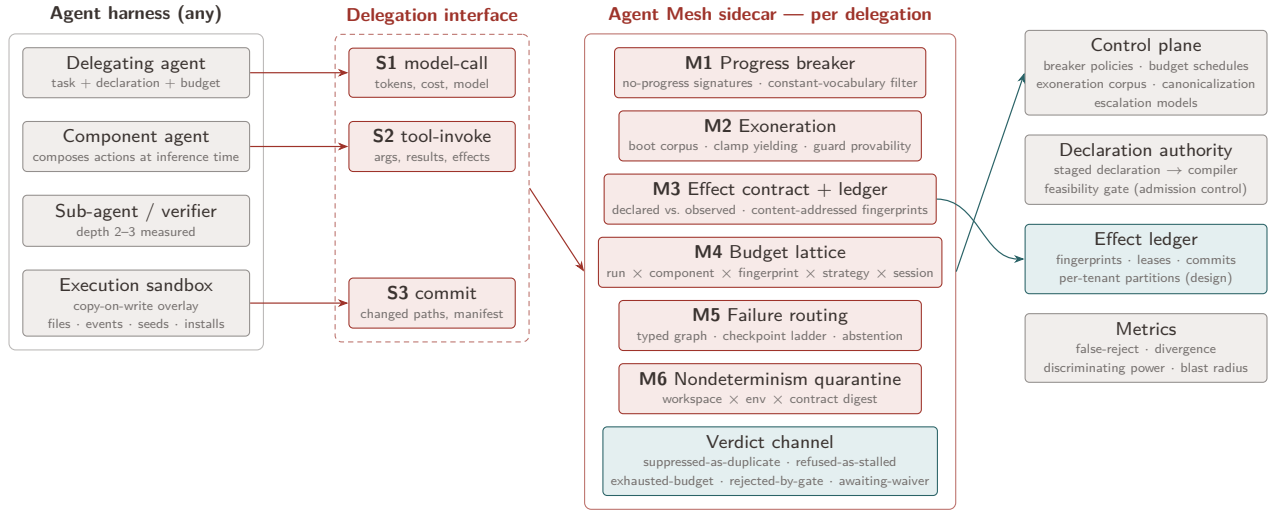
\begin{figure*}[t]
\centering
\begin{tikzpicture}[fig, node distance=3mm]

% --- harness column
\node[bxg, minimum width=30mm] (dag) {Delegating agent\\[-1pt]\textcolor{mink!70}{\tiny task + declaration + budget}};
\node[bxg, minimum width=30mm, below=of dag] (cag) {Component agent\\[-1pt]\textcolor{mink!70}{\tiny composes actions at inference time}};
\node[bxg, minimum width=30mm, below=of cag] (sag) {Sub-agent / verifier\\[-1pt]\textcolor{mink!70}{\tiny depth 2--3 measured}};
\node[bxg, minimum width=30mm, below=of sag] (sbx) {Execution sandbox\\[-1pt]\textcolor{mink!70}{\tiny copy-on-write overlay}\\[-1pt]\textcolor{mink!70}{\tiny files $\cdot$ events $\cdot$ seeds $\cdot$ installs}};
\node[grp, fit=(dag)(cag)(sag)(sbx), label={[ttl]above:Agent harness (any)}] (harness) {};

% --- seams
\node[bxr, right=13mm of dag, minimum width=22mm] (s1) {\textbf{S1} model-call\\[-1pt]\textcolor{mink!70}{\tiny tokens, cost, model}};
\node[bxr, right=13mm of cag, minimum width=22mm] (s2) {\textbf{S2} tool-invoke\\[-1pt]\textcolor{mink!70}{\tiny args, results, effects}};
\node[bxr, right=13mm of sbx, minimum width=22mm] (s3) {\textbf{S3} commit\\[-1pt]\textcolor{mink!70}{\tiny changed paths, manifest}};
\node[grpr, fit=(s1)(s2)(s3), label={[ttlr]above:Delegation interface}] (seams) {};

% --- sidecar
\node[bxr, right=13mm of s1, minimum width=40mm] (m1) {\textbf{M1} Progress breaker\\[-1pt]\textcolor{mink!70}{\tiny no-progress signatures $\cdot$ constant-vocabulary filter}};
\node[bxr, below=1.6mm of m1, minimum width=40mm] (m2) {\textbf{M2} Exoneration\\[-1pt]\textcolor{mink!70}{\tiny boot corpus $\cdot$ clamp yielding $\cdot$ guard provability}};
\node[bxr, below=1.6mm of m2, minimum width=40mm] (m3) {\textbf{M3} Effect contract + ledger\\[-1pt]\textcolor{mink!70}{\tiny declared vs.\ observed $\cdot$ content-addressed fingerprints}};
\node[bxr, below=1.6mm of m3, minimum width=40mm] (m4) {\textbf{M4} Budget lattice\\[-1pt]\textcolor{mink!70}{\tiny run $\times$ component $\times$ fingerprint $\times$ strategy $\times$ session}};
\node[bxr, below=1.6mm of m4, minimum width=40mm] (m5) {\textbf{M5} Failure routing\\[-1pt]\textcolor{mink!70}{\tiny typed graph $\cdot$ checkpoint ladder $\cdot$ abstention}};
\node[bxr, below=1.6mm of m5, minimum width=40mm] (m6) {\textbf{M6} Nondeterminism quarantine\\[-1pt]\textcolor{mink!70}{\tiny workspace $\times$ env $\times$ contract digest}};
\node[bxt, below=1.6mm of m6, minimum width=40mm] (vc) {Verdict channel\\[-1pt]\textcolor{mink!70}{\tiny suppressed-as-duplicate $\cdot$ refused-as-stalled}\\[-1pt]\textcolor{mink!70}{\tiny exhausted-budget $\cdot$ rejected-by-gate $\cdot$ awaiting-waiver}};
\node[grp, draw=mred!70, fit=(m1)(m2)(m3)(m4)(m5)(m6)(vc), label={[ttlr]above:Agent Mesh sidecar --- per delegation}] (mesh) {};

% --- control plane + ledger
\node[bxg, right=13mm of m1.east, anchor=west, minimum width=32mm] (cp) {Control plane\\[-1pt]\textcolor{mink!70}{\tiny breaker policies $\cdot$ budget schedules}\\[-1pt]\textcolor{mink!70}{\tiny exoneration corpus $\cdot$ canonicalization}\\[-1pt]\textcolor{mink!70}{\tiny escalation models}};
\node[bxg, below=2.5mm of cp, minimum width=32mm] (dec) {Declaration authority\\[-1pt]\textcolor{mink!70}{\tiny staged declaration $\rightarrow$ compiler}\\[-1pt]\textcolor{mink!70}{\tiny feasibility gate (admission control)}};
\node[bxt, below=2.5mm of dec, minimum width=32mm] (led) {Effect ledger\\[-1pt]\textcolor{mink!70}{\tiny fingerprints $\cdot$ leases $\cdot$ commits}\\[-1pt]\textcolor{mink!70}{\tiny per-tenant partitions (design)}};
\node[bxg, below=2.5mm of led, minimum width=32mm] (mx) {Metrics\\[-1pt]\textcolor{mink!70}{\tiny false-reject $\cdot$ divergence}\\[-1pt]\textcolor{mink!70}{\tiny discriminating power $\cdot$ blast radius}};

\draw[arr] (dag.east) -- (s1.west);
\draw[arr] (cag.east) -- (s2.west);
\draw[arr] (sbx.east) -- (s3.west);
\draw[arr] (seams.east) -- (mesh.west);
\draw[art] (mesh.east) -- (cp.west);
\draw[art] (m3.east) to[out=0,in=180] (led.west);
\end{tikzpicture}
\caption{Agent Mesh architecture. The mesh is defined against an abstract delegation interface with
three interception seams (S1 model-call, S2 tool-invocation, S3 commit); any orchestrator exposing
these seams can host the sidecar. The data plane runs seven primitives per delegation and reports
through a verdict channel whose outcomes are deliberately distinct---in particular
\emph{suppressed-as-duplicate} (the ledger working) must not be confused with
\emph{refused-as-stalled} (the breaker tripping). The declaration authority sits in the control
plane because it is admission control: it decides whether a delegation set may be created at all.}
\label{fig:arch}
\end{figure*}

\section{Method}\label{sec:method}

\paragraph{What counts as an incident.} An incident is a recorded failure of a run or a component
that was diagnosed to a cause and, in the majority of cases, closed by a change. Incidents were
recorded operationally as they were diagnosed, in a running implementation log, not gathered
retrospectively for this paper. Each carries a numbered identifier; 81 carry a distinct run
identifier of the form \run{devrun\_<hex>} that indexes the run's persisted record, event stream, and
workspace.

\paragraph{How costs were measured.} Costs are taken from the platform's own durable
records---persisted attempts, recovery leases, budget documents, structured failure envelopes---and
from its event stream, not from reconstruction. Where a cost is a count of agent turns or tool calls,
it is a count of persisted records. Where it is a duration, it is the interval between logged events.
Where a run's component-level outcome is reported (for example six of six components completing, then
three), it is taken from the scheduler's own summary lines.

\paragraph{Validation of causes.} The platform's guards are mutation-tested: a guard added in
response to an incident is required to fail when the condition it guards is reintroduced. In the
course of this work two guards were deleted because no mutation could make them fail, on the
principle that a guard that cannot fail is not evidence. Where a diagnosis is reported below as
confirmed, confirmation means the fix was reverted and the failure reproduced.

\paragraph{Prediction as a check on diagnosis.} For one incident the diagnosis was used to predict a
specific numeric outcome in advance of the next run---the corrected value of a failing
assertion---which then reproduced three times. We note this because it is the strongest form of
confirmation available in an observational setting, and because it distinguishes a diagnosis from a
narrative fitted after the fact.

\paragraph{Diagnoses that were withdrawn.} Several diagnoses recorded during the period were
subsequently disproved by measurement and are recorded as withdrawn rather than deleted: among them,
an attributed duplicate-publish defect that measurement showed did not exist (the producer was
correct), and a suspected absence of a retry mechanism that was in fact present and had executed
twice in the run under examination. We report this because a corpus with no withdrawn diagnoses
should not be trusted.

\paragraph{Relation to established failure-study method.} The design follows the production
failure-study tradition in systems research---most directly Yuan et al.~\cite{yuan}, who analysed
198 user-reported failures across five distributed data-intensive systems to derive testable
generalisations. Our corpus is comparable in size (147 incidents) and narrower in scope (one system),
and differs in one respect that cuts both ways: their failures were user-reported and independently
sampled, whereas ours were diagnosed by the team operating the platform. That yields deeper causal
detail---we hold the durable records, and we could revert fixes to confirm---at the cost of the
independence sampling provides. We treat that as the study's principal limitation
(\Cref{sec:threats}) rather than as a detail.

\paragraph{Threats to the measurement itself} are stated in \Cref{sec:threats}. The principal ones
are that the corpus is single-system, that incidents are self-diagnosed by the team that built the
platform, and that the record over-represents failures interesting enough to be written down.

\section{Findings}\label{sec:findings}

\begin{figure*}[t]
\centering
\begin{tikzpicture}[fig]
%% ---- left panel: nested breaker scopes
\node[bx, minimum width=20mm, minimum height=5mm] (ts) at (0,0)
  {Tool session\\[-1pt]\textcolor{mink!70}{\tiny 4 no-progress turns $\cdot$ 10 check runs}};
\node[bxr, below=1.5mm of ts, minimum width=44mm]
  (vig) {\textcolor{mred}{\tiny 54 successful identical calls in 11 min: invisible to every error-based guard}};
\node[grp, fit=(ts)(vig), inner sep=3pt, label={[sub]below:recovery strategy: durable lease $\cdot$ graded refund}] (g1) {};
\node[grp, fit=(g1), inner sep=6pt, label={[sub]below:failure fingerprint: max attempts per fingerprint}] (g2) {};
\node[grp, fit=(g2), inner sep=6pt, label={[sub]below:component delegation: graded no-progress stop}] (g3) {};
\node[grp, fit=(g3), inner sep=6pt, label={[sub]below:run: wall clock unlimited; per-check timeouts + patience}] (g4) {};
\node[ttl, above=1mm of g4] {Breaker scopes (nested)};

%% ---- right panel: signal adequacy
\node[bxg, right=26mm of g4.east, anchor=west, minimum width=54mm] (r1)
  {Round evidence\\[-1pt]\textcolor{mink!70}{\tiny identifiers reported by the failed attempt}};
\node[bxr, below=4mm of r1, minimum width=54mm] (r2)
  {Constant-vocabulary filter\\[-1pt]\textcolor{mink!70}{\tiny discard identifiers the strategy re-emits unchanged}\\[-1pt]\textcolor{mink!70}{\tiny (e.g.\ the failing check name); keep evidence that can vary}};
\node[bxg, below=4mm of r2, minimum width=54mm] (r3)
  {One shared progress function\\[-1pt]\textcolor{mink!70}{\tiny stall detector and budget refund read the same value}};
\node[bxt, below=4mm of r3, minimum width=54mm] (r4)
  {Trip / refund / escalate\\[-1pt]\textcolor{mteal}{\tiny discriminating power $=1-$ fraction decided on all-constant evidence}\\[-1pt]\textcolor{mteal}{\tiny (was 0 for four recovery paths)}};
\draw[ar] (r1) -- (r2); \draw[arr] (r2) -- (r3); \draw[ar] (r3) -- (r4);
\node[ttl, above=1mm of r1] {Signal adequacy (per decision)};
\end{tikzpicture}
\caption{Left: breaker scopes in the primary system, from run down to tool session, each with its own
signal and budget; the highlighted vignette is a loop made entirely of successful calls. Right: the
signal-adequacy pipeline. Round evidence is filtered against the delegation's constant vocabulary
before both progress guards read one shared function, so the stall detector and the budget refund
cannot disagree about what a round measured.}
\label{fig:breaker}
\end{figure*}
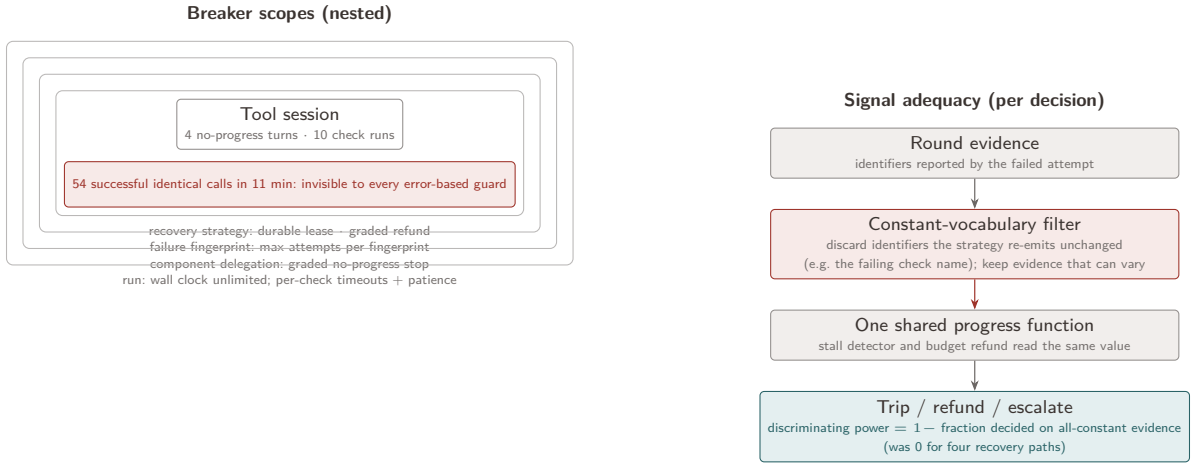

\subsection{F1: agents fail by ceasing to converge, not by erroring}

The fifty-four-call loop described in the introduction is not an outlier in kind. Across the corpus, the
dominant failure mode of a delegation is not an exception but a sequence of individually successful
actions that stops changing the outcome. This has a direct consequence for the inherited primitive:
an error-rate breaker observes nothing. In the verifier incident every call returned success, so
every error-based guard in the system---and there were several---was structurally blind. The loop
ended because a human noticed.

The platform now runs progress-based breakers at five nested scopes (\Cref{fig:breaker}, left):
four byte-identical tool calls, four session turns yielding no new grounding or mutation, a
per-fingerprint repeat bound, a per-component graded stop, and a run-level bound. Session progress
is content-keyed rather than path-keyed: a read returns the file's digest with its content, so
re-reading a file whose bytes changed is grounding while an identical re-read is not. That
distinction was itself forced by an incident in which post-write verification reads were being
counted as wandering.

\begin{figure}[t]
\centering
\begin{tikzpicture}[fig, x=13mm, y=4.2mm]
% axes
\draw[draw=mink!45] (0,0) -- (5.4,0);
\draw[draw=mink!45] (0,0) -- (0,6.4);
\foreach \y in {0,1,2,3,4,5,6} \node[sub, left=1mm] at (0,\y) {\y};
\node[sub, rotate=90, left=5mm] at (0,3.2) {failing tests};
\foreach \x/\l in {1/r1,2/r2,3/r3} \node[sub, below=1mm] at (\x,0) {\l};
\node[sub, below=4mm] at (2,0) {repair rounds inside the killed window};

% the trajectory: crash -> 5 failing -> 2 failing (3 newly passing)
\draw[arr, line width=0.7pt] (1,6) -- (2,5) -- (3,2);
\fill[mred] (1,6) circle (1.1pt);
\fill[mred] (2,5) circle (1.1pt);
\fill[mred] (3,2) circle (1.1pt);
\node[sub, above right=0mm and 1mm] at (1,6) {collection crash};
\node[sub, above right=0mm and 1mm] at (2,5) {5 failing};
\node[sub, right=1.5mm] at (3,2) {\textbf{2 failing}, 3 newly passing};

% the breaker verdict
\draw[draw=mred, dash pattern=on 2pt off 1.5pt] (3,-0.4) -- (3,6.4);
\node[bxr, right=3mm of {(3,4.6)}, anchor=west, align=left, minimum width=26mm]
 {\textcolor{mred}{\tiny breaker verdict at r3:}\\[-1pt]\textcolor{mred}{\tiny \textbf{``evidence has not moved''}}};

% the constant the guard actually hashed
\node[bx, below=8mm of {(2.6,0)}, anchor=north, minimum width=68mm, align=left]
 {\textcolor{mink!70}{\tiny What both guards hashed each round: \texttt{["stock-integration"]} --- the failing}\\[-1pt]
  \textcolor{mink!70}{\tiny \emph{check name}. Constant by construction, so the stall detector saw a fixed}\\[-1pt]
  \textcolor{mink!70}{\tiny hash and the graded refund saw a fixed count of one. Two guards, one blind spot.}};
\end{tikzpicture}
\caption{Signal adequacy, measured. Inside the window that killed the leading component the evidence
moved monotonically toward green---one missing dictionary key from passing---while the breaker
declared it unmoved, because the identifier it fingerprinted was a property of the strategy rather
than of the round.}
\label{fig:movement}
\end{figure}
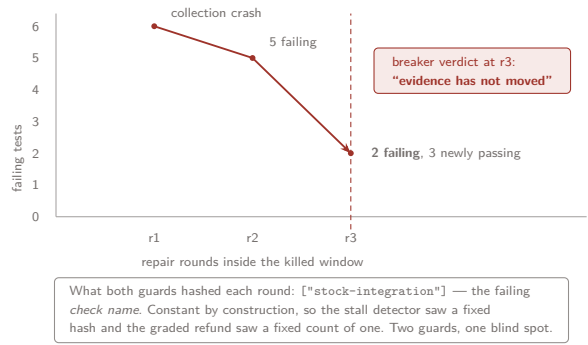

\subsection{F2: a progress signal can be constant by construction}

Progress-based breaking replaces one failure mode with another. Our stall detector fingerprinted the
identifiers a failed attempt reported. For one whole class of failure those identifiers were the
\emph{failing check name}---a property of the recovery strategy, identical whether the agent had
fixed four defects or none.

Both progress guards read that constant. The stall detector saw a fixed hash and tripped on the
third round regardless of progress; the graded budget refund, which exists precisely to stop
count-based ceilings killing converging loops, saw a fixed count of one and could never fire. Every
model-driven repair in the system was therefore guaranteed to be declared stalled on its third
round.

The measured cost: one run peaked at six of six components with cross-service acceptance executing
and ended at three. Neither step down was a model defect. Inside the window that killed the leading
component the evidence had moved from a collection crash, to five failing tests, to two failing with
three newly passing---both survivors a single missing dictionary key from green (\Cref{fig:movement}).

Two further recovery paths were found to fingerprint on constants by replay---one on a failure-class
enumeration value, one on check names---and would have failed identically on their third round.

\emph{The finding generalizes.} A no-progress signal computed over identifiers that are constant by
construction is not a conservative breaker; it is a guaranteed false trip. Signal adequacy must be
demonstrated, not assumed.

\begin{figure}[t]
\centering
\begin{tikzpicture}[fig, x=7.2mm, y=1.55mm]
\draw[draw=mink!45] (0,0) -- (7.2,0);
\draw[draw=mink!45] (0,0) -- (0,24);
\foreach \y in {0,6,12,18,21} \node[sub, left=1mm] at (0,\y) {\y};
\node[sub, rotate=90, left=5mm] at (0,12) {events in the log};
\foreach \x in {1,...,6} \node[sub, below=1mm] at (\x,0) {i\x};
\node[sub, below=4mm] at (3.5,0) {check invocations of the same delegation};

% cumulative 3 per invocation
\foreach \x/\y in {1/3,2/6,3/9,4/12,5/15,6/21}{\fill[mteal] (\x,\y) circle (1.1pt);}
\draw[art, line width=0.7pt] (1,3) -- (2,6) -- (3,9) -- (4,12) -- (5,15) -- (6,21);

% the assertion line
\draw[draw=mred, dash pattern=on 2pt off 1.5pt] (0,1) -- (7.2,1);
\node[sub, right=0.5mm, text=mred] at (5.6,2.6) {test asserts \textbf{1}};
\node[sub, right=1mm, text=mteal] at (6.05,21) {\textbf{21}};

\node[bx, below=8mm of {(3.6,0)}, anchor=north, minimum width=68mm, align=left]
 {\textcolor{mink!70}{\tiny The producer was correct: \emph{exactly three events per invocation}, six times.}\\[-1pt]
  \textcolor{mred}{\tiny The duplication was in the ledger, not the producer.}\\[-1pt]
  \textcolor{mink!70}{\tiny The log lived outside the workspace so the cleanliness check would not revert}\\[-1pt]
  \textcolor{mink!70}{\tiny it --- and therefore outside everything that cleans. 35 such logs on disk.}};
\end{tikzpicture}
\caption{A measured duplicate-effect harm. Effects committed by previous invocations of the same
delegation remained visible to the current one, so a correct, idempotent component became
unwinnable. This is the boundary transactional containment does not reach.}
\label{fig:eventlog}
\end{figure}
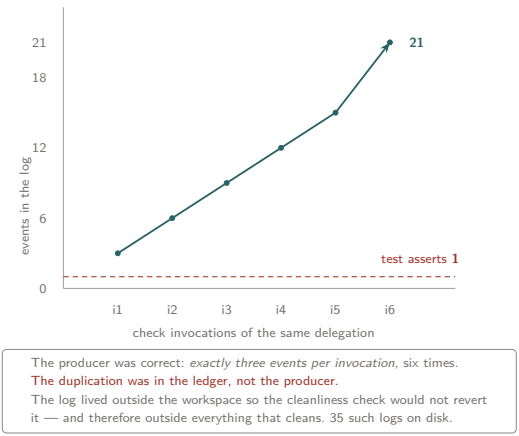

\subsection{F3: effects outlive the delegations that commit them}

The platform contains transactional containment: mutating sessions run in a copy-on-write overlay
whose commit refuses any changed path outside a declared writable set. Containment works, and for
the effect class it contains a crashed or retried delegation commits nothing twice.

It does not contain everything. The event-log incident above occurred at a boundary deliberately
placed outside the workspace, and therefore outside every mechanism that cleans it: 21 events across
six invocations, 35 such logs on disk, the oldest four days old (\Cref{fig:eventlog}). The producer
was correct. The asymmetry that named the cause is that the acceptance harness truncated its log
every test while the per-service path never did.

Four further effect classes escape containment and are undeduplicated: package-registry resolution
and installation, per-service database writes, post-write hook execution, and model-provider calls.
The last is the most expensive and the least visible---provider retries spend tokens with no ledger
of what was already spent.

\emph{One effect class in the platform is deduplicated}, and its design is instructive: recovery
actions acquire a durable lease keyed on failure, strategy, strategy version, and budget key, with a
unique-index violation as the deduplication signal and orphan reconciliation on orchestrator restart.
This is the developer-enumerated-key design, working exactly for the class someone thought to
enumerate---and the measured duplicate-effect failure occurred in a class nobody did.

\begin{figure}[t]
\centering
\begin{tikzpicture}[fig, node distance=3.2mm]
\node[bxt, minimum width=30mm] (c1) {command accepted};
\node[bxt, below=of c1, minimum width=30mm] (c2) {event published};
\node[bxt, below=of c2, minimum width=30mm] (c3) {subscriber received};
\node[bxt, below=of c3, minimum width=30mm] (c4) {state committed};
\node[bxt, below=of c4, minimum width=30mm] (c5) {route read};
\draw[art] (c1) -- (c2); \draw[art] (c2) -- (c3); \draw[art] (c3) -- (c4); \draw[art] (c4) -- (c5);
\node[sub, right=3mm of c1.east, anchor=west, align=left] {absent $\rightarrow$ publisher / transport};
\node[sub, right=3mm of c2.east, anchor=west, align=left] {absent $\rightarrow$ transport / subscriber lifecycle};
\node[sub, right=3mm of c3.east, anchor=west, align=left] {absent $\rightarrow$ consumer / repository};
\node[sub, right=3mm of c4.east, anchor=west, align=left] {present but stale $\rightarrow$ transform / commit};
\node[sub, right=3mm of c5.east, anchor=west, align=left] {bad read $\rightarrow$ route / query binding};
\node[ttlt, above=2mm of c1] {The first missing checkpoint localizes the transition};
\node[bxr, below=5mm of c5, minimum width=74mm, align=left]
 {\textcolor{mred}{\tiny One correlation id per test, propagated across five process boundaries.}\\[-1pt]
  \textcolor{mink!70}{\tiny Bounded metadata only --- component, stage, channel, operation. Never payloads.}\\[-1pt]
  \textcolor{mink!70}{\tiny Ambiguous evidence $\rightarrow$ abstain and fall back, never exclude the true owner.}};
\end{tikzpicture}
\caption{Failure routing by checkpoint ladder. Spans are effect transitions rather than RPC calls;
the owner of the first unproven transition is the routing target.}
\label{fig:ladder}
\end{figure}

\subsection{F4: misrouted failure attribution damages correct work}

In a fleet where delegations repair one another, attributing a failure to the wrong delegation is not
a wasted retry. It is a mandate handed to a correct component to edit code that was already right.

Acceptance recovery originally mapped a failing test to its declared scenario identifiers and
reopened every dependency whose artifacts declared one of them. That rule conflates three distinct
facts: that a component has tests covering a scenario, that it participates in the scenario's runtime
path, and that it owns the transition that failed. In one incident a failure caused by two components
woke five, and \textbf{three bystanders regressed working code}. In another, a failing assertion
polling one service's endpoint was routed to two components that merely declared the scenario, while
the service that owned the stale state was never nominated.

A second cost is diagnostic rather than destructive. When repair briefs for cross-process failures
nominated only unmodifiable files---a frozen test, a platform conftest, a third-party plugin---one
repair window consumed \textbf{943 tool turns, 451 of them read or search operations, across 71
minutes}: 48\% of the effort spent re-deriving causality the platform had already computed and
discarded.

\begin{table*}[t]
\centering
\small
\renewcommand{\arraystretch}{1.25}
\begin{tabular}{@{}p{6.0cm}p{4.4cm}p{5.6cm}@{}}
\toprule
\textbf{what the layer blocked} & \textbf{measured cost} & \textbf{why it was wrong} \\
\midrule
Repair clamp restricted reads \emph{and} writes to a mis-diagnosed target set &
107 turns, \textbf{zero accepted writes}, component lost &
The one file holding the defect was outside the diagnosis; the delegation could not even read it \\
Husk check rejected a framework's documented base-class idiom &
16 rejected writes across 4 services &
Each rejection costs a turn against the no-progress budget \\
Seed gate matched a handler by callee name &
3 corrections + 2 stronger-model rounds, component lost &
The platform's \emph{own} generated dispatch seam read as ``nothing is seeded'' \\
Appeal verified a citation against only the file the model named &
3 real quoted lines refused &
Every frozen source is shown to the model; all must therefore be citable \\
Appeal \emph{accepted} against a platform-generated artifact &
Component killed holding a correct diagnosis &
A wrongly-permissive verdict forecloses every remaining route \\
Stall detector fingerprinted a constant identifier &
Run 6/6 $\rightarrow$ 3/6 &
Every model-driven repair guaranteed stalled on round three \\
Scenario coverage used as causal identity &
Blast radius 5; \textbf{3 bystanders regressed working code} &
Coverage is not causation; the true owner was never nominated \\
Preview oracle recognized only bearer tokens &
6 loops, $\sim$1 hour, byte-identical signature &
Services authenticate via gateway-injected headers; repair could not satisfy it either way \\
Provenance walls demanded ratchet-only artifacts &
Fully-green verifier-approved run failed &
A freeze record and an \emph{optional} cache entry required in a mode that produces neither \\
Seed gate unioned every dependency's fixtures &
11 unsatisfiable identifiers; recovery budget burned &
Deriving rather than copying the set collapsed it to 3 real rows \\
Toolchain defaults left unstated (async mode; fixture dry run) &
6 and 8 escalated corrections &
Failures named only plugin internals; no edit the model could make would help \\
Stray zero-byte file from a test runner &
Byte-identical evidence until terminal &
A subprocess artifact was indistinguishable from an undeclared agent write \\
\bottomrule
\end{tabular}
\caption{Enforcement-layer failures: the layer blocking \emph{correct} work. Each row is a distinct
production incident with a measured cost. This failure mode, not permitting a forbidden action, is
the characteristic failure of an enforcement layer for agent delegation.}
\label{tab:exoneration}
\end{table*}

\subsection{F5: the enforcement layer is a primary source of outages}

This is the finding we did not anticipate and consider the most transferable.

When an enforcement layer rejects work that is in fact correct, the agent complies, is rejected
again, produces byte-identical evidence, and burns its entire budget against a wall. The delegation
is \emph{unwinnable}, and every reliability primitive above it is measuring a fiction. We recorded
twelve distinct instances (\Cref{tab:exoneration}). The most expensive cost 107 agent turns and
\emph{zero accepted writes}, because a repair clamp restricted both writes and reads to a
mis-diagnosed target set, so the delegation could not even read the file that held its defect.

Two sub-patterns are worth separating. First, several gates blocked correct work because they encoded
one mode's assumptions---a shape with direct analogues in distributed systems. Zhang et al.~\cite{upgrade} show that upgrade
failures arise when a component's assumptions and its environment diverge, and Yin et
al.~\cite{misconfig} find that a majority of misconfigurations are parameter mistakes that violate a
rule the system itself holds --- an assumption encoded in a checker rather than in the checked. Ours differ in that the
diverging assumption belongs to the \emph{checker} rather than to the system under change: a gate rejected a framework's own documented base-class idiom sixteen times
across four services; another read the platform's \emph{own} generated dispatch seam as ``nothing is
seeded'' because it matched handlers by callee name.

Second, and less obvious, \emph{a wrongly-permissive enforcement decision can be worse than a
wrongly-restrictive one}. An appeal mechanism intended as the escape hatch resolved a correct
diagnosis against the wrong file and \emph{accepted} it, returning a verdict that foreclosed every
remaining repair route. A rejection leaves the delegation a path; a mistaken acceptance does not.

\subsection{F6: some effects are observable only when the whole system runs}

Effects exist that no tool-boundary, sandbox, or per-service check can reach: cross-service
authentication, an empty database, a navigation route linked but never generated. Booting the
generated application---every backend service as a real out-of-process server, the frontend as a real
dev server---and probing it as a gateway and a browser would, caught three escapes that every other
boundary had passed: a frontend shipping a complete design system with no pipeline to compile it (a
green build, an unstyled application), and a multi-page navigation wired over routes that were never
generated (a green build, a 404 on prefetch).

The same boundary immediately produced its own instance of F5. A run reached this gate fully green
and verifier-approved, then looped \textbf{six times over roughly an hour on one finding with a
byte-identical signature}. The probe recognized only bearer tokens; the platform's scaffolded services
authenticate via gateway-injected identity headers and declare no security scheme because
authentication lives upstream. The probe called without those headers, the service \emph{correctly}
returned 401, and the finding was unwinnable by repair---declaring security in the service merely
inverted it. A new evidence boundary is also a new surface on which correct work can be blocked.

\subsection{F7: a stable oracle cannot be assumed}

Consistent checkpointing of a distributed computation is well understood~\cite{chandylamport}, and
stream processors achieve exactly-once state by combining it with deterministic replay --- Carbone et
al.'s asynchronous barrier snapshotting~\cite{abs} persists operator state at consistent cuts and
replays records from the cut on recovery. The platform's resume path relies on the same idea: a
delegation's durable checkpoint is only meaningful if the state it names can be reconstructed.

What both approaches assume, and what agent delegation violates, is that re-executing from a
checkpoint against unchanged inputs yields the same outcome. A dataflow operator replayed over the
same records is deterministic by construction; a delegation replayed over the same workspace composes
its actions afresh at inference time, and its oracle is a suite the delegation itself authored.
Replay therefore recovers position but not behaviour, which is why the platform quarantines rather
than retries.

Microservice retry assumes that the same request against the same state yields the same verdict, so a
differing result is information. For agent delegation the oracle is a test suite the agent itself
authored. A flaky oracle makes every primitive above it lie: the breaker trips on noise, a ledger
would fingerprint a non-reproducible effect, the router attributes a phantom. The platform keys every
check observation on a triple---a content digest of the workspace, a digest of the resolved execution
environment, and a digest of the check contract---and quarantines the evidence when two observations
under one key disagree, rather than retrying.

\subsection{Corpus statistics and measurement scope}\label{sec:stats}

\paragraph{What the corpus supports, and what it does not.} The incident record is a chronological
operational log in which incidents are numbered and cross-referenced, not a structured database with
a category field. A category distribution over all 147 incidents would therefore have to be assigned
retrospectively by the same people who diagnosed them, and we do not report one: a distribution
produced that way would measure our labelling more than the system. What we report instead are
quantities that were recorded mechanically at the time---workspace contents, persisted run records,
and the event stream---together with the per-incident costs of the subset the paper analyses
directly.

For the same reason, no inferential statistics are reported. The corpus is a single system's
operational record, not a sample from a population, and incidents were neither randomly selected nor
independently observed; significance testing against it would be a category error. Descriptive
statistics with explicit $N$ and ranges are the strongest claim the design supports, and the
controlled evaluation of \Cref{sec:eval} is what would license anything stronger.

\paragraph{Workload scale ($N=15$ archived runs).} Median 78 source files and 2{,}231 lines of Python
and TypeScript per run; range 4--110 files and 507--12{,}240 lines; maximum two backend services plus
a frontend. \textbf{Three of the fifteen produced no source at all}---runs that terminated before any
component committed work. That failure-severity rate ($3/15$) is itself a measurement, and it is the
one figure here that generalises least: it reflects the platform's state during a period of active
change rather than a steady-state defect rate.

\paragraph{Enforcement-layer incidents ($N=12$).} The costs in \Cref{tab:exoneration} are recorded in
heterogeneous units---agent turns, rejected writes, repair loops, failed tests---because the incidents
terminate at different stages, and we do not aggregate across them. Of the four measured in agent
turns, the range is 107 turns (zero accepted writes) to 943 turns (451 of them read or search
operations). Of the twelve, \emph{all} ended in either a component's terminal failure or an exhausted
recovery budget on work that was subsequently confirmed correct.

\subsection{Workload and a fully-traced run}\label{sec:measured}

One run is recorded in full, and is reported here because it shows the machinery converging rather
than failing.

\paragraph{A run traced end to end.} One run (\run{devrun\_228979e8}) is recorded in full: four hours
fifty-five minutes, five components, ten scheduler cycles across two attempts. Its persisted record
contains 12 terminal component failures, 23 stall verdicts, 3 escalated rounds granted by the graded
stop, and 3 checkpoints discarded as stale on re-entry. Nine consecutive scheduler cycles in the
first attempt ended with the cross-service acceptance component failing. The second attempt reached
\emph{21 of 21 required checks passing}, and the independent verifier---which had returned
\code{approved=False} fourteen minutes earlier---returned \code{approved=True} with zero findings.

We report this run because it shows the machinery working as designed rather than the failures the
rest of the paper documents: a rejection followed by repair followed by approval, with the stall
detector firing 23 times without terminating a component that was still converging, and the
escalation path spending three stronger-model rounds rather than three funerals.

\section{The cross-cutting finding: identity adequacy and evidence adequacy}\label{sec:spine}

Five subsystem failures in the corpus share a cause that is not visible from any one of them
(\Cref{tab:spine}): an identity that failed to discriminate produced a confident wrong answer. A
progress fingerprint over a check name; a commit identifier keyed on a transaction rather than on
content; a graph node named for a logical projection, collapsing two physically distinct service
databases; scenario coverage used as causal identity; and a work planner measuring coupling across
components rather than per component.

\paragraph{Evidence adequacy, the dual.} The same corpus yields a second requirement that is not a
restatement of the first. Identity adequacy asks whether a signal can \emph{distinguish} two states
that differ; evidence adequacy asks whether the signal can \emph{change at all}, and whether it is
entitled to be acted upon. Four instances recur. A stop may fire only on evidence capable of moving
(F2): a fingerprint over identifiers constant by construction is not conservative, it is a guaranteed
false trip. A router must \emph{abstain} when evidence is ambiguous rather than confidently exclude
the true owner (F4). A mutation kill counts only when it is \emph{attributable} to the rule it
targets---otherwise a mutant that anchored on load-bearing code reports safety that was never
demonstrated. And a check may be trusted only when its outcome is deterministic under identical
workspace and environment conditions (F7), since a flaky oracle makes every primitive above it lie.

The two halves fail differently and must be checked separately. An inadequate identity produces a
confident wrong answer; inadequate evidence produces a confident answer to a question that was never
measured. Both were present in our corpus, and in the breaker they were present simultaneously---the
same constant identifier defeated the stall detector and the graded refund at once, which is why a
single fix repaired both and why we now require both guards to read one function.

\begin{table*}[t]
\centering
\small
\renewcommand{\arraystretch}{1.3}
\begin{tabular}{@{}p{3.2cm}p{8.2cm}p{5.4cm}@{}}
\toprule
\textbf{subsystem} & \textbf{identity that failed to discriminate} & \textbf{consequence} \\
\midrule
circuit breaker & progress fingerprint over the failing \emph{check name}---a property of the
strategy, identical whether four defects were fixed or none & every model-driven repair guaranteed
stalled on round three; the run went 6/6 $\rightarrow$ 3/6 \\
effect ledger & commit id keyed on \emph{transaction id}, so identical content from two transactions
hashes differently & cannot deduplicate; the fingerprint is provenance, not identity \\
topology graph & a logical node name collapsing \emph{two different service databases} into one
projection & false edge, false green graph \\
failure attribution & \emph{scenario coverage} used as causal identity & five components woken for a
two-component fault; three bystanders regressed working code \\
work planner & coupling measured \emph{across} components rather than per component & two components
asserting one scenario would have had their artifact sets welded into a single slice \\
\bottomrule
\end{tabular}
\caption{One failure, five subsystems: an identity that does not discriminate produces a confident
wrong answer. The last two rows derived the same rule independently, for different objects.}
\label{tab:spine}
\end{table*}

\paragraph{The rule was derived twice, independently.} The strongest evidence that this is a property
of the problem rather than a habit of one team is that two subsystems arrived at it separately, for
different objects, without inheriting it from each other. The topology graph refuses to collapse
identically-shaped state living in different processes. The work planner, written for a different
purpose by a different path, refuses to collapse a scenario that two components both assert---its
coupling is measured per component, so a shared scenario forms a cluster in each of them and never
welds their artifact sets together. Both arrived there after the naive version produced a confident
wrong answer.

\section{Primitives implied by the findings}\label{sec:primitives}

The findings imply a set of reliability primitives whose enforcement unit is the \emph{delegation}
rather than the message: the logical task that carries a declaration, holds a budget, acquires a
lease, commits effects, and can be retried or resumed as a unit. \Cref{fig:arch} situates them
against three interception seams---the model-call boundary, the tool-invocation boundary, and the
commit boundary---so the design is stated against an interface rather than against our architecture.

\paragraph{P1: progress-based breaking under a signal-adequacy obligation (F1, F2).} Trip on
no-progress signatures rather than error rate, and compute the signal against the delegation's
\emph{constant vocabulary}---the identifiers a strategy re-emits unchanged by construction---
discarding evidence drawn entirely from it. Both progress guards must read one function so they
cannot disagree about what a round measured. The corresponding metric is \emph{discriminating
power}: the fraction of breaker decisions taken on evidence not wholly constant. Before the fix it
was zero for four recovery paths.

\begin{figure*}[t]
\centering
\begin{tikzpicture}[fig, node distance=7mm]
\node[bxg, minimum width=24mm] (d1) {\textbf{1. Delegate}\\[-1pt]\textcolor{mink!70}{\tiny task + budget + declaration}};
\node[bxg, right=of d1, minimum width=24mm] (d2) {\textbf{2. Execute}\\[-1pt]\textcolor{mink!70}{\tiny overlay sandbox; actions at inference}};
\node[bxr, right=of d2, minimum width=24mm] (d3) {\textbf{3. Observe}\\[-1pt]\textcolor{mink!70}{\tiny tool + sandbox boundary}};
\node[bxr, right=of d3, minimum width=24mm] (d4) {\textbf{4. Verify}\\[-1pt]\textcolor{mink!70}{\tiny observed vs.\ declared; divergence is the signal}};
\node[bxt, right=of d4, minimum width=24mm] (d5) {\textbf{5. Commit}\\[-1pt]\textcolor{mink!70}{\tiny fingerprint effects $\rightarrow$ ledger}};
\draw[ar] (d1) -- (d2); \draw[ar] (d2) -- (d3); \draw[ar] (d3) -- (d4); \draw[ar] (d4) -- (d5);

\node[bxr, below=10mm of d2, minimum width=26mm] (flt) {Fault\\[-1pt]\textcolor{mink!70}{\tiny crash $\cdot$ timeout $\cdot$ restart $\cdot$ partition}};
\node[bxg, right=10mm of flt, minimum width=26mm] (rty) {Retry / resume\\[-1pt]\textcolor{mink!70}{\tiny same logical delegation}};
\node[bxt, right=10mm of rty, minimum width=34mm] (chk) {Effect-ledger check\\[-1pt]\textcolor{mink!70}{\tiny fingerprint present? yes $\rightarrow$ suppress / reconcile}\\[-1pt]\textcolor{mink!70}{\tiny no $\rightarrow$ execute + record}\\[-1pt]\textcolor{mteal}{\tiny verdict: suppressed-as-duplicate}};
\draw[ard] (d2) -- (flt);
\draw[arr] (flt) -- (rty);
\draw[art] (rty) -- (chk);
\draw[art] (chk.east) -- ++(4mm,0) |- (d5.east);

\node[bx, draw=mink!35, below=6mm of flt, minimum width=76mm, xshift=22mm, align=center]
  (naive) {\textcolor{mink!60}{\tiny naive mesh retry (no ledger): re-executes committed effects
  $\rightarrow$ duplicate events, seeds, installs, provider calls}};
\draw[ard] (flt.south) |- (naive.west);
\end{tikzpicture}
\caption{Delegation lifecycle under the effect contract. Effects observed at the tool and sandbox
boundaries are verified against the declaration and fingerprinted into the ledger at commit. A
retried or resumed delegation deduplicates at the effect boundary; the
\emph{suppressed-as-duplicate} verdict is what distinguishes correct deduplication from a breaker
trip. The dashed path is the naive mesh policy the baseline arm measures.}
\label{fig:lifecycle}
\end{figure*}
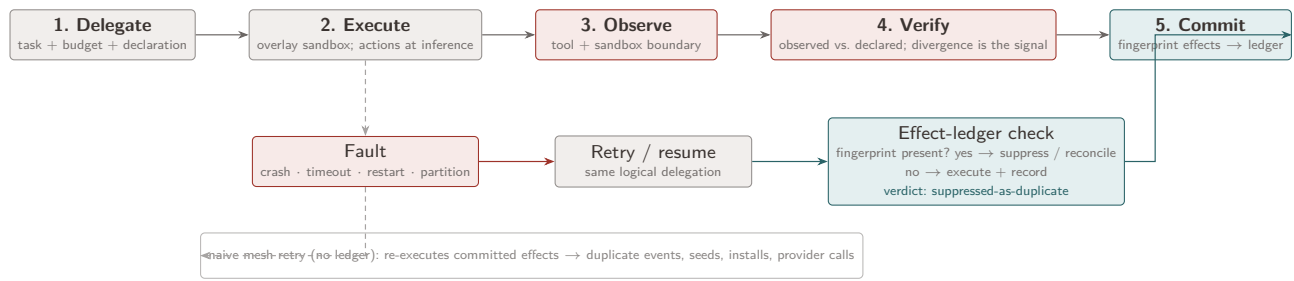

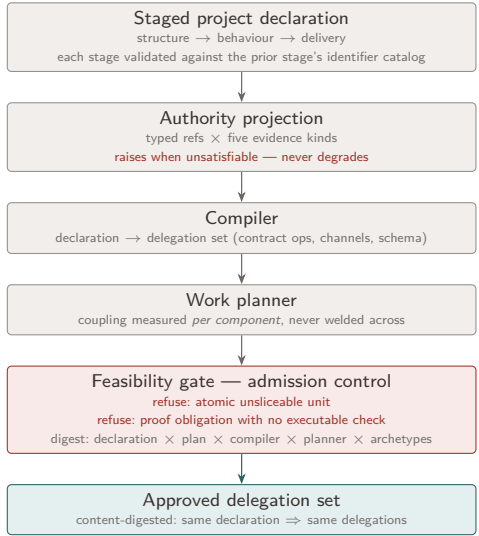
\begin{figure}[t]
\centering
\begin{tikzpicture}[fig, node distance=4mm]
\node[bxg, minimum width=62mm] (stg) {Staged project declaration\\[-1pt]\textcolor{mink!70}{\tiny structure $\rightarrow$ behaviour $\rightarrow$ delivery}\\[-1pt]\textcolor{mink!70}{\tiny each stage validated against the prior stage's identifier catalog}};
\node[bxg, below=of stg, minimum width=62mm] (aut) {Authority projection\\[-1pt]\textcolor{mink!70}{\tiny typed refs $\times$ five evidence kinds}\\[-1pt]\textcolor{mred}{\tiny raises when unsatisfiable --- never degrades}};
\node[bxg, below=of aut, minimum width=62mm] (cmp) {Compiler\\[-1pt]\textcolor{mink!70}{\tiny declaration $\rightarrow$ delegation set (contract ops, channels, schema)}};
\node[bxg, below=of cmp, minimum width=62mm] (pln) {Work planner\\[-1pt]\textcolor{mink!70}{\tiny coupling measured \emph{per component}, never welded across}};
\node[bxr, below=of pln, minimum width=62mm] (fea) {Feasibility gate --- admission control\\[-1pt]\textcolor{mred}{\tiny refuse: atomic unsliceable unit}\\[-1pt]\textcolor{mred}{\tiny refuse: proof obligation with no executable check}\\[-1pt]\textcolor{mink!70}{\tiny digest: declaration $\times$ plan $\times$ compiler $\times$ planner $\times$ archetypes}};
\node[bxt, below=of fea, minimum width=62mm] (run) {Approved delegation set\\[-1pt]\textcolor{mink!70}{\tiny content-digested: same declaration $\Rightarrow$ same delegations}};
\draw[ar] (stg) -- (aut); \draw[ar] (aut) -- (cmp); \draw[ar] (cmp) -- (pln);
\draw[ar] (pln) -- (fea); \draw[art] (fea) -- (run);
\node[sub, right=2mm of fea.east, anchor=west, text width=0mm] {};
\end{tikzpicture}
\caption{The declaration authority. Declarations are compiled, not hand-written per delegation, and
approval is gated by a dry-run compile that refuses a declaration the platform cannot build. This is
enforcement-layer exoneration at declaration altitude: prove the layer can be satisfied before
committing work to it.}
\label{fig:declaration}
\end{figure}

\paragraph{P2: the effect contract (F3).} Declaration alone is a comment; observation alone has no
reference. The mesh is a hybrid: declarations are compiled rather than hand-authored
(\Cref{fig:declaration}), and committed effects are observed at boundaries requiring no agent
cooperation, with divergence as the enforcement signal (\Cref{fig:lifecycle}). Observation is
mechanical here only because the tool surface is closed; \emph{enumerability of the surface, not of
the effect set, is the enabling property}.

\paragraph{P3: the effect ledger (F3).} A fingerprint over committed effects---canonicalized tool
calls, argument digests, external mutation identifiers---against which a retried or resumed
delegation deduplicates. This is the one primitive that is \emph{specified but not built}, and we
mark it as such throughout. The delta is small and named: the platform's transaction already returns
a commit identifier over transaction, component, phase and content manifest, and dropping the
transaction identifier makes it content-addressed. Nothing in the study demonstrates this primitive
works; the study demonstrates the harm it addresses.

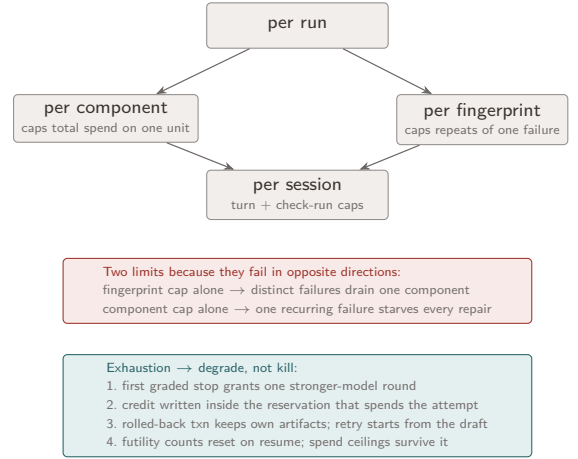
\begin{figure}[t]
\centering
\begin{tikzpicture}[fig, node distance=4mm]
\node[bxg, minimum width=24mm] (run) {per run};
\node[bxg, below left=6mm and 1mm of run, minimum width=22mm] (comp) {per component\\[-1pt]\textcolor{mink!70}{\tiny caps total spend on one unit}};
\node[bxg, below right=6mm and 1mm of run, minimum width=22mm] (fp) {per fingerprint\\[-1pt]\textcolor{mink!70}{\tiny caps repeats of one failure}};
\node[bxg, below=16mm of run, minimum width=24mm] (sess) {per session\\[-1pt]\textcolor{mink!70}{\tiny turn + check-run caps}};
\draw[ar] (run) -- (comp); \draw[ar] (run) -- (fp);
\draw[ar] (comp) -- (sess); \draw[ar] (fp) -- (sess);
\node[bxr, below=5mm of sess, minimum width=62mm, align=left]
 (why) {\textcolor{mred}{\tiny Two limits because they fail in opposite directions:}\\[-1pt]
        \textcolor{mink!70}{\tiny fingerprint cap alone $\rightarrow$ distinct failures drain one component}\\[-1pt]
        \textcolor{mink!70}{\tiny component cap alone $\rightarrow$ one recurring failure starves every repair}};
\node[bxt, below=4mm of why, minimum width=62mm, align=left]
 (deg) {\textcolor{mteal}{\tiny Exhaustion $\rightarrow$ degrade, not kill:}\\[-1pt]
        \textcolor{mink!70}{\tiny 1. first graded stop grants one stronger-model round}\\[-1pt]
        \textcolor{mink!70}{\tiny 2. credit written inside the reservation that spends the attempt}\\[-1pt]
        \textcolor{mink!70}{\tiny 3. rolled-back txn keeps own artifacts; retry starts from the draft}\\[-1pt]
        \textcolor{mink!70}{\tiny 4. futility counts reset on resume; spend ceilings survive it}};
\end{tikzpicture}
\caption{The budget scope lattice, and the degradation ladder that replaces wall-clock kill. Budgets
attach per run, component, failure fingerprint, strategy, and session; the per-fingerprint and
per-component limits are separate quantities because they fail in opposite directions.}
\label{fig:budget}
\end{figure}

\paragraph{P4: budget attenuation over a scope lattice, degrading rather than killing (F1).} Budgets
key per run, per component, per failure-evidence fingerprint, per recovery strategy, and per session
(\Cref{fig:budget}). Per-fingerprint and per-component limits are separate quantities because they
fail in opposite directions: a fingerprint cap alone lets distinct failures drain one component; a
component cap alone lets one recurring failure starve every other repair. Exhaustion degrades---the
first graded stop grants one round on a stronger model, with the credit written inside the same
atomic reservation that consumes the attempt---rather than killing on a clock.

\paragraph{P5: failure routing (F4).} A typed topology graph with scoped node identities and
evidence-graded edges, plus a five-stage checkpoint ladder carried by a correlation identifier across
process boundaries, in which \emph{the first missing checkpoint localizes the transition}
(\Cref{fig:ladder}). This is distributed tracing whose spans are effect transitions rather than
remote calls. Abstention is a first-class outcome: ambiguous evidence falls back rather than
confidently excluding the true owner. Structure is necessary and never sufficient---a green graph
never marks acceptance passed.

\paragraph{P6: enforcement-layer exoneration (F5, F6).} The layer must be proven not to block correct
traffic, at runtime rather than by review. Three mechanisms: boot-time exoneration, where the service
refuses to start if any gate rejects a member of a corpus of independently-verified-correct artifacts
(thirteen gates, $\sim$36\,ms, each proven armed by mutating its corpus artifact);
clamp yielding, where an enforcement scope derived from a diagnosis admits a file refused twice, on
the reasoning that a delegation which owns a file and keeps naming it is telling you the diagnosis
was wrong; and guard provability, where a guard no mutation can make fail is deleted. Refusal
semantics are part of the contract: \emph{suppressed-as-duplicate}, \emph{refused-as-stalled},
\emph{exhausted-budget}, and \emph{rejected-by-gate} must be distinct observable verdicts, because in
our own system all three of the first conditions collapsed into one fatal outcome and the
duplicate-lease case is the deduplication mechanism \emph{working correctly}.

\paragraph{P7: nondeterminism quarantine (F7).} Key observations on workspace, environment, and
contract digests, and quarantine rather than retry when two observations under one key disagree.

\section{Deployment outcomes}\label{sec:outcomes}

\begin{table*}[t]
\centering
\small
\renewcommand{\arraystretch}{1.25}
\begin{tabular}{@{}p{3.4cm}p{2.5cm}p{10.6cm}@{}}
\toprule
\textbf{mechanism} & \textbf{status} & \textbf{evidence, or what remains} \\
\midrule
Effect declaration + transparent verification & running, measured &
Overlay commit refuses undeclared writes; agent/subprocess writes discriminated by declaration.
Divergence rate computable today and backfillable from archived workspaces. \\
Declaration authority + feasibility gate & running &
Staged declaration, deterministic compile, dry-run admission control; refuses unsliceable units and
obligations with no executable check. \\
Progress breaker + signal adequacy & running, measured &
Constant-vocabulary filter; both guards read one function. Pre-fix discriminating power was
\emph{zero} for four recovery paths. \\
Budget lattice + degradation & running, measured &
Graded refund on strict improvement; unlimited wall clock; one stronger-model round before the stop
(four of eleven strategies wired). Per-delegation token ledger \emph{not built}. \\
Failure routing (graph + ladder) & running; success criteria unvalidated &
Blast radius 5\,$\rightarrow$\,2 measured. Two-point graph build and checkpoint ladder running;
the motivating routing case is \emph{not yet observed live}. \\
Enforcement exoneration & running, measured &
Thirteen gates checked at boot; clamp yielding fired five times (previously structurally zero);
unprovable guards deleted. \\
Nondeterminism quarantine & running &
Workspace $\times$ environment $\times$ contract digest; conflicting outcomes quarantined. \\
Running-system verification & running, measured &
Preview oracle over the booted multi-service app; three escapes caught that no other boundary
reaches. Browser-behaviour oracle \emph{designed, unbuilt}. \\
\midrule
Effect ledger (content-addressed fingerprint) & \textbf{specified delta} &
One field dropped from an existing commit identifier; the manifest function already exists. One
measured duplicate-effect incident motivates it. \emph{Not built.} \\
Distinct refusal verdicts & \textbf{specified delta} &
Admission control currently collapses stall, exhaustion, and duplicate-lease into one outcome.
Required before breaker precision/recall means anything. \emph{Not built.} \\
Per-tenant ledger partitioning & \textbf{designed, unbuilt} &
The deployed platform is single-tenant; the tenancy model is evaluated on the reference harness
only. \\
Effect trace store; token/cost ledger & \textbf{designed, unbuilt} &
Prerequisites for every Phase~1 and Phase~2 number. The application log is not a substitute. \\
\bottomrule
\end{tabular}
\caption{Mechanism status. The upper block is running in production; the lower block is what the
evaluation still requires. We state this as a table because the paper's central claim about itself
is that the distinction is never blurred.}
\label{tab:status}
\end{table*}

\Cref{tab:status} records what is running, what is specified, and what is designed. Where a primitive
was deployed, the observed effect was:

\begin{itemize}
\item \textbf{Signal adequacy (P1).} The constant-vocabulary filter made repeated-evidence stops
rare rather than routine; the converging replay that previously died on round three now proceeds,
and the stuck replay still stops. Verified by reverting in both directions.
\item \textbf{Clamp yielding (P6).} Five admissions in one run, including the exact file that had
killed a component the previous day. Before the change the count was structurally zero.
\item \textbf{Failure routing (P5).} Blast radius fell from five components to two on the
arrangement case.
\item \textbf{Running-system verification (P6, F6).} Three escapes caught that no other boundary
reaches, each subsequently closed at its own layer.
\item \textbf{Declaration admission control (P2).} Two services complied with a newly-declared
physical schema contract on the first attempt, with no correction round, where the previous
undeclared arrangement had cost seven of eight acceptance tests twice from two different causes.
\end{itemize}

The platform now carries a project end to end to a live browser preview, and the run traced in
\Cref{sec:measured} completed fully green. On the test-driven substrate the best acceptance result
remains seven of eight tests passing.

\subsection{Ablation: removing the verification ladder}\label{sec:ablation}

The platform supports an explicit, labelled \emph{direct implementation mode} that removes the
test-driven ratchets---cluster slicing, test preflight, oracle qualification, red validation,
mutation qualification, and freeze---for a component too large to slice. Runs in this mode are the
first that reached a live browser preview, which makes them an ablation of the verification ladder
conducted in production rather than in a harness (\Cref{tab:ablation}).

The design is quasi-experimental rather than randomised: the mode is selected for architectural
reasons, not assigned, so the two arms differ in workload as well as in treatment. What it does
establish is \emph{which defects each rung was absorbing}, because every defect in
\Cref{tab:ablation} surfaced only once its rung was removed, and each was subsequently closed at a
layer that operates in both modes.

\begin{table*}[t]
\centering
\small
\renewcommand{\arraystretch}{1.25}
\begin{tabular}{@{}p{3.5cm}p{6.4cm}p{6.1cm}@{}}
\toprule
\textbf{rung removed} & \textbf{defect that surfaced} & \textbf{measured cost, and closure} \\
\midrule
Cluster slicing & An atomic component the planner could not slice was admitted without a size gate &
Convergence-complexity 120 against a budget of 6; closed by making the mode an explicit flag the
feasibility gate consults \\
Oracle qualification and seed gating & Required-fixture set derived as a raw union over every
dependency & 11 unsatisfiable identifiers; recovery budget exhausted on a correct component. Closed
by \emph{deriving} the set rather than copying it: 11 $\rightarrow$ 3 real rows \\
Freeze and provenance & Final verifier demanded a freeze record, and separately an optional cache
entry, for every component & A fully-green, verifier-approved run failed. Closed by scoping the
provenance requirement to modes that produce the artifact \\
(none --- new boundary) & Preview oracle recognised only bearer tokens while services authenticate
via gateway-injected headers & 6 repair loops, $\sim$1 hour, byte-identical signature; unwinnable by
repair. Closed by making the oracle authenticate as the gateway does \\
\bottomrule
\end{tabular}
\caption{Ablation of the verification ladder. Each defect surfaced only when its rung was removed,
and each was closed at a layer that operates in both modes---so the ladder's contribution is
diagnostic rather than merely procedural. The final row is the dual result: a \emph{new} evidence
boundary introduced its own instance of F5.}
\label{tab:ablation}
\end{table*}

Two smaller ablations validate individual mechanisms by reversion rather than by removal of a whole
stage. The constant-vocabulary filter of F2 was verified in both directions: with the rule reverted,
a converging replay is declared stalled on its third round; with the diagnostics fallback reverted,
a genuinely stuck replay never stops. Guard provability is applied the same way as a standing
policy---a guard that no mutation can make fail is deleted, and two were removed on those grounds
during the period.

\paragraph{What the ablation does and does not show.} It does not show the ladder is necessary for
delivery: the ablated runs reached a live preview, which the non-ablated ones had not. It shows what
each rung was \emph{holding up}---every defect it exposed is an identity-or-evidence defect of the
shape \Cref{sec:spine} predicts, and each had been absorbed silently rather than reported. The
ablation was therefore productive in the diagnostic sense, and it is the reason the last-mile
mechanisms it forced into existence (preview oracle, seed derivation, mode-scoped provenance) apply
in both modes.

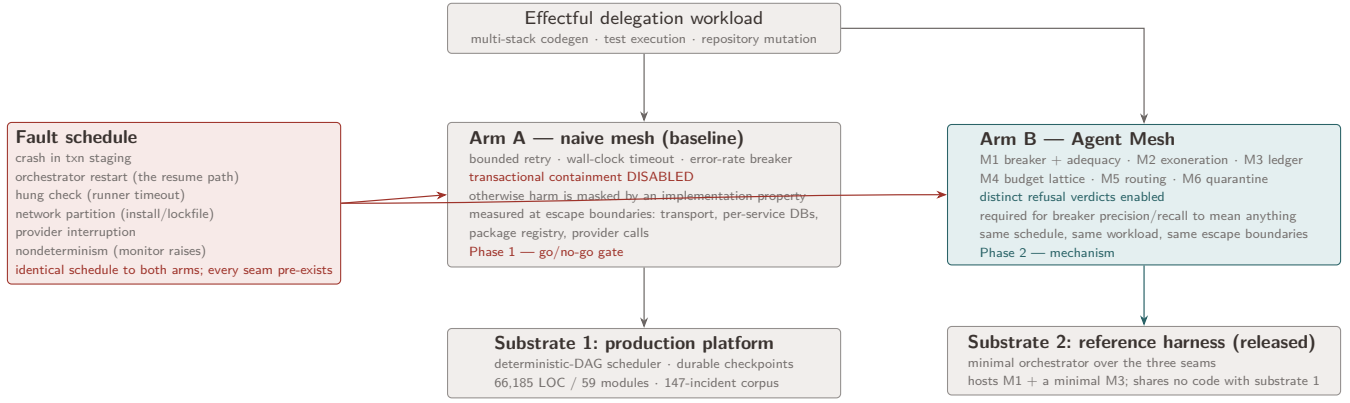
\begin{figure*}[t]
\centering
\begin{tikzpicture}[fig, node distance=6mm]
\node[bxg, minimum width=52mm] (wl) {Effectful delegation workload\\[-1pt]\textcolor{mink!70}{\tiny multi-stack codegen $\cdot$ test execution $\cdot$ repository mutation}};

\node[bxr, below left=9mm and 14mm of wl, minimum width=44mm, align=left] (fs)
 {\textbf{Fault schedule}\\[-1pt]\textcolor{mink!70}{\tiny crash in txn staging}\\[-1pt]\textcolor{mink!70}{\tiny orchestrator restart (the resume path)}\\[-1pt]\textcolor{mink!70}{\tiny hung check (runner timeout)}\\[-1pt]\textcolor{mink!70}{\tiny network partition (install/lockfile)}\\[-1pt]\textcolor{mink!70}{\tiny provider interruption}\\[-1pt]\textcolor{mink!70}{\tiny nondeterminism (monitor raises)}\\[-1pt]\textcolor{mred}{\tiny identical schedule to both arms; every seam pre-exists}};

\node[bxg, below=9mm of wl, minimum width=52mm, align=left] (a) {\textbf{Arm A --- naive mesh (baseline)}\\[-1pt]\textcolor{mink!70}{\tiny bounded retry $\cdot$ wall-clock timeout $\cdot$ error-rate breaker}\\[-1pt]\textcolor{mred}{\tiny transactional containment DISABLED}\\[-1pt]\textcolor{mink!70}{\tiny otherwise harm is masked by an implementation property}\\[-1pt]\textcolor{mink!70}{\tiny measured at escape boundaries: transport, per-service DBs,}\\[-1pt]\textcolor{mink!70}{\tiny package registry, provider calls}\\[-1pt]\textcolor{mred}{\tiny Phase 1 --- go/no-go gate}};

\node[bxt, right=14mm of a, minimum width=52mm, align=left] (b) {\textbf{Arm B --- Agent Mesh}\\[-1pt]\textcolor{mink!70}{\tiny M1 breaker + adequacy $\cdot$ M2 exoneration $\cdot$ M3 ledger}\\[-1pt]\textcolor{mink!70}{\tiny M4 budget lattice $\cdot$ M5 routing $\cdot$ M6 quarantine}\\[-1pt]\textcolor{mteal}{\tiny distinct refusal verdicts enabled}\\[-1pt]\textcolor{mink!70}{\tiny required for breaker precision/recall to mean anything}\\[-1pt]\textcolor{mink!70}{\tiny same schedule, same workload, same escape boundaries}\\[-1pt]\textcolor{mteal}{\tiny Phase 2 --- mechanism}};

\node[bxg, below=8mm of a, minimum width=52mm, align=left] (s1) {\textbf{Substrate 1: production platform}\\[-1pt]\textcolor{mink!70}{\tiny deterministic-DAG scheduler $\cdot$ durable checkpoints}\\[-1pt]\textcolor{mink!70}{\tiny 66{,}185 LOC / 59 modules $\cdot$ 147-incident corpus}};
\node[bxg, below=8mm of b, minimum width=52mm, align=left] (s2) {\textbf{Substrate 2: reference harness (released)}\\[-1pt]\textcolor{mink!70}{\tiny minimal orchestrator over the three seams}\\[-1pt]\textcolor{mink!70}{\tiny hosts M1 + a minimal M3; shares no code with substrate 1}};

\draw[ar] (wl) -- (a);
\draw[ar] (wl.east) -| (b);
\draw[arr] (fs.east) -- (a.west);
\draw[arr] (fs.east) to[out=0,in=180] (b.west);
\draw[ar] (a) -- (s1); \draw[art] (b) -- (s2);
\end{tikzpicture}
\caption{Evaluation design. One workload and one fault schedule drive two arms: the baseline runs
naive service-mesh policies with transactional containment disabled---without which the harm number
measures a property of our implementation rather than of agent delegation---and the mesh arm enables
the seven primitives with distinct refusal verdicts.}
\label{fig:eval}
\end{figure*}

\section{The controlled evaluation this study motivates}\label{sec:eval}

This study is observational. The controlled evaluation it motivates is a two-arm fault-injection
design (\Cref{fig:eval}) over the same workload: a baseline arm applying naive service-mesh policies
with \emph{transactional containment disabled}---without which the harm number measures a property of
our implementation rather than of agent delegation---and a mesh arm enabling the primitives with
distinct refusal verdicts.

Co-primary metrics are duplicate-effect rate at the boundaries containment does not reach, and work
destroyed per fault. Two measurements from this study bound the second in advance: the 943-turn
repair window of F4, and the 107-turn zero-write component of F5. If neither primary metric is
material under realistic fault injection, the premise is weak and the work stops---a designed kill
criterion.

\begin{table}[t]
\centering
\small
\renewcommand{\arraystretch}{1.2}
\begin{tabular}{@{}p{4.3cm}p{3.5cm}@{}}
\toprule
\textbf{metric} & \textbf{status} \\
\midrule
\multicolumn{2}{@{}l}{\emph{Obtainable from the existing record}} \\
Delegation depth distribution & \textbf{reported} (1/8/7 at depths 1/2/3) \\
Discriminating power, pre-fix & \textbf{reported} (0 for four paths) \\
Declaration--observation divergence & backfillable from archived workspaces \\
Enforcement incidents & \textbf{reported} (\Cref{tab:exoneration}) \\
Blast radius, before/after & \textbf{reported} (5 $\rightarrow$ 2) \\
\midrule
\multicolumn{2}{@{}l}{\emph{Blocked on an instrument}} \\
Work destroyed per fault & needs token/cost ledger \\
Tokens discarded by timeout kills & needs token ledger + baseline arm \\
Duplicate-effect rate (mesh arm) & needs fingerprint + effect trace \\
Fingerprint collision / miss & needs the fingerprint \\
Breaker precision/recall & needs distinct refusal verdicts \\
Sidecar overhead (p50/p99) & no sidecar boundary exists yet \\
Ledger storage per tenant & no ledger; no tenancy \\
\midrule
\multicolumn{2}{@{}l}{\emph{Blocked on the experiment}} \\
Duplicate-effect rate (baseline) & needs the containment-disabled arm \\
Task success at fixed budget & needs both arms to complete the workload \\
\bottomrule
\end{tabular}
\caption{Metric inventory. Five quantities are already derivable from the operating record; the rest
are gated on instruments or on the two-arm study, and we say which.}
\label{tab:metrics}
\end{table}

\Cref{tab:metrics} separates the metrics already derivable from the operating record from those
gated on an instrument or on the study itself. The instruments not yet built are stated plainly: a
per-delegation token and cost ledger; an effect trace as a persisted store rather than the
application log; the content-addressed effect fingerprint; sidecar observation of network
destinations and spawned processes; distinct refusal verdicts; and per-tenant ledger partitioning,
which the single-tenant deployed platform does not implement.

\section{Threats to validity}\label{sec:threats}

\begin{description}[leftmargin=0em,itemsep=3pt]
\item[Observational, not controlled.] This is the study's defining limitation and we state it
first. Incidents were observed, not induced; no baseline arm ran; and the deployment outcomes of
\Cref{sec:outcomes} are before/after observations on a system that was changing for other reasons at
the same time. Confounding is therefore possible for every one of them, and none should be read as an
effect size.

We claim the genre rather than the guarantee. Production failure studies are an established
instrument in systems research precisely because some failure classes appear only at deployment
scale: Yuan et al.~\cite{yuan} derive testable generalisations from 198 observed failures without a
controlled arm, and Zhang et al.~\cite{upgrade} do the same for upgrade failures. What such studies
license is the identification and characterisation of failure classes, not measurement of how much a
proposed remedy helps. That boundary governs what we assert: the seven findings are claims about
\emph{what goes wrong and why}, each traceable to recorded incidents; the primitives of
\Cref{sec:primitives} are the response those findings imply, and their effectiveness is
\emph{unmeasured}. \Cref{sec:eval} specifies the two-arm design that would measure it, with a
designed kill criterion, and \Cref{tab:metrics} states which quantities are already derivable and
which are gated on instruments that do not yet exist. \Cref{sec:ablation} is the one quasi-
experimental element, and it is quasi-experimental rather than controlled because the treatment is
selected for architectural reasons rather than assigned.
\item[Single system, self-diagnosed.] One platform, diagnosed by the team that built it. Mitigations:
causes were confirmed by reverting the fix and reproducing the failure wherever the paper says
``confirmed''; guards are mutation-tested and two were deleted for being unprovable; withdrawn
diagnoses are recorded rather than removed. None of that substitutes for an independent replication,
which the released fault schedules and reference harness are intended to enable.
\item[Survivorship in the corpus.] The record over-represents failures interesting enough to write
down. Routine failures that were fixed without comment are under-represented, so frequencies should
not be inferred from it. Costs attached to individual incidents are measured; the corpus's
composition is not a sample.
\item[The effect ledger is unbuilt.] P3 is specified, not demonstrated. The study establishes the
harm it addresses---a measured duplicate-effect failure at a boundary containment does not
reach---and specifies the delta. It does not show the primitive works.
\item[Strict-platform bias.] F5 is partly a consequence of how much this platform enforces. A permissive orchestrator has fewer gates and therefore fewer
opportunities to block correct work---and correspondingly weaker guarantees. We think the trade is
general and the failure mode under-reported, but the frequencies we observed are ours.
\item[Prior-art engagement.] The related-work positioning below is stated at the level of research
lanes. Specific systems are cited only where we could resolve them to a verified record; several
named in an earlier draft were removed rather than cited from memory. This is a deliberate
under-citation, and closing it is the first task before any venue submission.
\end{description}

\section{Positioning and related work}\label{sec:related}

Two lanes of agent-infrastructure work are adjacent to this study and neither asks its question. The
\emph{authorization} lane---capability metadata attached to values, information-flow labelling,
control-flow-integrity checking of agent invocations---governs whether an action \emph{should be
permitted}, under an adversarial model of prompt injection. The \emph{transport} lane---generative-AI
deframing proxies, Model Context Protocol~\cite{mcp} interceptors, agent-to-agent gateways built on
sidecar infrastructure~\cite{istio,linkerd,envoy}---enforces on JSON-RPC syntax. Neither asks whether
an authorized action has already been committed, whether a delegation is still converging, or whether
the enforcement layer itself is wrong.

\subsection{Durable execution and exactly-once semantics}

The nearest prior art is durable workflow execution~\cite{durablefn,temporal,cadence,durablefunctions},
whose semantics are formalised by Burckhardt et al.~\cite{durablefn}: durable state is reconstructed
by deterministic replay against a log of externally-observed effects, and exactly-once activity
semantics rest on developer-supplied idempotency keys at \emph{enumerated} activity boundaries. The distinction this study sharpens is empirical rather than
argued: our platform contains that design and we can show where it stops. Recovery actions, the one
class enumerated in advance, are deduplicated by a durable lease and it works; the measured
duplicate-effect failure of F3 occurred in a class nobody enumerated, because an agent generates its
effectful operations at inference time and there is no site at which the key could have been
attached.

\begin{table*}[t]
\centering
\small
\renewcommand{\arraystretch}{1.25}
\begin{tabular}{@{}p{3.5cm}ccccc@{}}
\toprule
& \textbf{effect-level} & \textbf{progress-based} & \textbf{failure} & \textbf{enforcement} &
\textbf{unit of} \\
\textbf{approach} & \textbf{dedup} & \textbf{breaking} & \textbf{attribution} & \textbf{correctness}
& \textbf{enforcement} \\
\midrule
Service mesh \cite{istio,linkerd,envoy} & --- (key required) & --- (error rate) & --- & --- & request \\
Durable execution \cite{durablefn,temporal,cadence} & enumerated boundaries & --- (retry policy) & --- & --- & activity \\
Agent authorization \cite{camel,fides} & --- & --- & --- & partial\,$^{a}$ & action \\
Transport proxies \cite{mcp} & --- & --- & --- & --- & message \\
Distributed tracing \cite{dapper} & --- & --- & spans (RPC) & --- & call \\
MAS failure taxonomy \cite{mast} & --- & --- & task-level & --- & task \\
\midrule
\textbf{Agent Mesh (this work)} & \textbf{inference-time} & \textbf{signal-adequate} &
\textbf{effect transitions} & \textbf{boot + runtime} & \textbf{delegation} \\
\bottomrule
\end{tabular}
\caption{Capability comparison. ``---'' denotes not provided by the approach, not a deficiency:
each row solves a different problem well. The gap this paper addresses is the empty
\emph{effect-level deduplication} column for operation sets generated at inference time, and the
empty \emph{enforcement correctness} column entirely.
$^{a}$Authorization systems verify that a policy is enforced, not that the enforcement layer admits
correct traffic.}
\label{tab:sota}
\end{table*}

\Cref{tab:sota} places this work against the approaches an agent orchestrator would otherwise reach
for. The comparison is capability-based rather than quantitative because no shared benchmark exists:
the systems compared do not accept the same workload, and constructing one is the subject of
\Cref{sec:eval} rather than of this study. Two columns are empty for every prior approach.
\emph{Effect-level deduplication} is provided only where the effectful operation set is enumerated in
advance, which agent delegation precludes by construction. \emph{Enforcement correctness}---whether
the layer can be shown not to block correct traffic---is, as far as our survey extends, claimed by no
existing system in either lane.

\subsection{Agent orchestration frameworks}

Multi-agent orchestration frameworks~\cite{autogen} compose agents that converse, delegate, and
invoke tools, typically over the reasoning-and-acting loop formalised by Yao et
al.~\cite{react}, with tool invocation itself the subject of a line of work from
Toolformer~\cite{toolformer} onward. They provide the delegation structure this paper's findings concern, and their
reliability affordances are those of ordinary application code: retry on exception, a step or
recursion ceiling, and a timeout. None of the findings here is a criticism of a particular
framework---F1 through F7 are stated against any hierarchical agent--subagent orchestration exposing
the three seams of \Cref{fig:arch}, and we deliberately avoid claiming a framework-specific result
we did not measure.

\subsection{Empirical failure studies}

The closest prior work is empirical rather than architectural. Cemri et al.~\cite{mast} construct
a failure taxonomy for multi-agent LLM systems from over 200 tasks across seven frameworks,
identifying fourteen failure modes in three categories: specification issues, inter-agent
misalignment, and task verification. Our study is complementary and differs on three axes. Theirs is
\emph{cross-framework and task-level}, ours is \emph{single-system and infrastructure-level}: the
failures we report are not failures of agents reasoning or coordinating but of the reliability
machinery around them---a breaker tripping on a constant, a ledger outliving its delegation, an
enforcement gate blocking correct work. Theirs is annotated from traces by external raters; ours is
recorded operationally with costs taken from the platform's own durable records and causes confirmed
by reverting fixes. And where MAST asks why a multi-agent system produces a wrong answer, we ask why
a correct agent is prevented from producing a right one---the unwinnable-delegation class of F5,
which a task-level taxonomy does not surface because the agent's own behaviour is not at fault.

\subsection{Automated program repair}

The delegations studied here repair code, which places the work adjacent to automated program
repair~\cite{apr,aprbib}. That field asks how to \emph{generate} a correct patch given a failing
test, and evaluates repair techniques by the correctness of the patches they produce. Our concern is
upstream of the patch and orthogonal to its quality: whether the repair loop is allowed to run at
all, whether the failure was routed to the delegation that owns it, whether the evidence the loop
stops on can move, and whether the effects of a repeated attempt are committed twice. A repair
technique that generates perfect patches still fails if the enforcement layer refuses its writes
(F5), if the fault is attributed to a component that cannot fix it (F4), or if the loop is declared
stalled while converging (F2). The two literatures compose: repair supplies the patch, the mesh
supplies the conditions under which attempting one is safe and terminating.

\subsection{Tracing, attribution, and evaluation}

Relative to distributed tracing~\cite{dapper,otel}, the contribution implied by F4 is spans that are
effect transitions rather than remote calls. Dapper established the span-and-trace model and
OpenTelemetry~\cite{otel} standardised it, but a span in both records that a \emph{call} happened;
the checkpoint ladder records whether an \emph{effect} was committed, which is what a routing
decision needs and what a call-level span cannot supply. The distinction matters for a delegation
whose failure mode is an absent write rather than a failed request. Benchmarks for autonomous software-engineering agents~\cite{swebench} and for agent capability more
broadly~\cite{agentbench} measure task resolution under controlled conditions, and Yehudai et
al.~\cite{agenteval} survey the field, noting that cost-efficiency, safety and robustness remain
under-assessed relative to capability. This study sits in that gap on the reliability side:
benchmarks measure task resolution; this study measures what the surrounding orchestrator must do for such an agent to be
retried, resumed, and repaired safely at all.

Finally, the findings are stated against \emph{any} hierarchical agent--subagent orchestration, not
against a particular framework. Nothing in F1--F7 depends on how delegations are expressed; they
depend only on a delegation being effectful, generating its operation set at inference time, costing
tokens whether or not its work is kept, and being retried, resumed, or repaired by a peer. Any
orchestrator with the three seams of \Cref{fig:arch} exhibits the same surface.

\section{Conclusion}

We studied 147 recorded failures in a production agentic delivery platform and found that the three
assumptions service-mesh reliability rests on---idempotence, latency as the failure signal, and free
discards---are each violated, with measured consequences: a fifty-four-call loop invisible to every
error-based guard, a progress signal that was constant by construction and drove a run from six of
six components to three, twenty-one events surviving across six invocations of one delegation to make
a correct component unwinnable, an attribution rule that woke five components for a two-component
fault and left three regressing working code, and twelve incidents in which the enforcement layer
blocked correct work.

The cross-cutting result is that five otherwise unrelated subsystems failed the same way---an
identity that did not discriminate, producing a confident wrong answer---and that two of them derived
the corrective rule independently, which is our best evidence that it is a property of the problem
rather than an artifact of one team. The primitives we derive follow from that: reliability for agent
delegation requires identities that discriminate and evidence that can move, and an enforcement layer
must be proven at runtime not to block correct work.

What this study does not do is compare against a controlled baseline. \Cref{sec:eval} specifies that
evaluation, with a designed kill criterion, and states plainly which instruments must exist before it
can produce numbers.

\paragraph{Availability.} Supplementary material accompanying this preprint documents the platform's
evidence boundaries, recovery machinery, declaration pipeline, and incident corpus, with an explicit
statement of what is not built. The fault schedules, workloads, and framework-neutral reference
harness of \Cref{sec:eval} are released with the controlled evaluation.


\begin{thebibliography}{9}
\bibitem{camel} E. Debenedetti, I. Shumailov, T. Fan, J. Hayes, N. Carlini, D. Fabian, C. Kern,
C. Shi, A. Terzis, and F. Tram\`er, ``Defeating prompt injections by design,''
arXiv:2503.18813, 2025.
\bibitem{fides} M. Costa, B. K\"opf, A. Kolluri, A. Paverd, M. Russinovich, A. Salem, S. Tople,
L. Wutschitz, and S. Zanella-B\'eguelin, ``Securing AI agents with information-flow control,''
arXiv:2505.23643, 2025.
\bibitem{agenteval} A. Yehudai, L. Eden, A. Li, G. Uziel, Y. Zhao, R. Bar-Haim, A. Cohan, and
M. Shmueli-Scheuer, ``Survey on evaluation of LLM-based agents,'' arXiv:2503.16416, 2025.
\bibitem{abs} P. Carbone, G. F\'ora, S. Ewen, S. Haridi, and K. Tzoumas, ``Lightweight asynchronous
snapshots for distributed dataflows,'' arXiv:1506.08603, 2015.
\bibitem{agentbench} X. Liu, H. Yu, H. Zhang, et al., ``AgentBench: evaluating LLMs as agents,''
arXiv:2308.03688, 2023.
\bibitem{misconfig} Z. Yin, X. Ma, J. Zheng, Y. Zhou, L. N. Bairavasundaram, and S. Pasupathy,
``An empirical study on configuration errors in commercial and open source systems,'' in \emph{Proc.\
23rd ACM Symp.\ Operating Systems Principles (SOSP)}, 2011.
\bibitem{apr} C. Le Goues, M. Pradel, and A. Roychoudhury, ``Automated program repair,''
\emph{Commun.\ ACM}, vol.~62, no.~12, pp.~56--65, 2019.
\bibitem{aprbib} M. Monperrus, ``Automatic software repair: a bibliography,'' \emph{ACM Comput.\
Surv.}, vol.~51, no.~1, pp.~1--24, 2018.
\bibitem{toolformer} T. Schick, J. Dwivedi-Yu, R. Dess\`i, R. Raileanu, M. Lomeli, L. Zettlemoyer,
N. Cancedda, and T. Scialom, ``Toolformer: language models can teach themselves to use tools,''
arXiv:2302.04761, 2023.
\bibitem{otel} OpenTelemetry Authors, ``OpenTelemetry: an observability framework and toolkit,''
Cloud Native Computing Foundation, \url{https://opentelemetry.io}. Accessed 2026.
\bibitem{autogen} Q. Wu et al., ``AutoGen: enabling next-gen LLM applications via multi-agent
conversation,'' arXiv:2308.08155, 2023.
\bibitem{chandylamport} K. M. Chandy and L. Lamport, ``Distributed snapshots: determining global
states of distributed systems,'' \emph{ACM Trans.\ Comput.\ Syst.}, vol.~3, no.~1, pp.~63--75, 1985.
\bibitem{react} S. Yao, J. Zhao, D. Yu, N. Du, I. Shafran, K. Narasimhan, and Y. Cao,
``ReAct: synergizing reasoning and acting in language models,'' in \emph{Proc.\ Int.\ Conf.\ Learning
Representations (ICLR)}, 2023; arXiv:2210.03629.
\bibitem{upgrade} Y. Zhang, J. Yang, Z. Jin, U. Sethi, K. Rodrigues, S. Lu, and D. Yuan,
``Understanding and detecting software upgrade failures in distributed systems,'' in \emph{Proc.\
28th ACM Symp.\ Operating Systems Principles (SOSP)}, 2021.
\bibitem{yuan} D. Yuan, Y. Luo, X. Zhuang, G. R. Rodrigues, X. Zhao, Y. Zhang, P. U. Jain, and
M. Stumm, ``Simple testing can prevent most critical failures: an analysis of production failures in
distributed data-intensive systems,'' in \emph{Proc.\ 11th USENIX Symp.\ Operating Systems Design and
Implementation (OSDI)}, 2014.
\bibitem{durablefn} S. Burckhardt, C. Gillum, D. Justo, K. Kallas, C. McMahon, and C. Meiklejohn,
``Durable functions: semantics for stateful serverless,'' \emph{Proc.\ ACM Program.\ Lang.}, vol.~5,
no.~OOPSLA, art.~133, 2021.
\bibitem{mast} M. Cemri, M. Z. Pan, S. Yang, et al., ``Why do multi-agent LLM systems fail?,''
arXiv:2503.13657, 2025.
\bibitem{swebench} C. E. Jimenez, J. Yang, A. Wettig, S. Yao, K. Pei, O. Press, and K. Narasimhan,
``SWE-bench: can language models resolve real-world GitHub issues?,'' arXiv:2310.06770, 2023.
\bibitem{dapper} B. H. Sigelman, L. A. Barroso, M. Burrows, P. Stephenson, M. Plakal, D. Beaver,
S. Jaspan, and C. Shanbhag, ``Dapper, a large-scale distributed systems tracing infrastructure,''
Google Technical Report, 2010.
\bibitem{istio} Istio Authors, ``Istio: connect, secure, control, and observe services,''
\url{https://istio.io}. Accessed 2026.
\bibitem{linkerd} Linkerd Authors, ``Linkerd: a service mesh for Kubernetes,''
\url{https://linkerd.io}. Accessed 2026.
\bibitem{envoy} Envoy Project Authors, ``Envoy proxy,''
\url{https://www.envoyproxy.io}. Accessed 2026.
\bibitem{mcp} Model Context Protocol, ``Specification,''
\url{https://modelcontextprotocol.io}. Accessed 2026.
\bibitem{temporal} Temporal Technologies, ``Temporal: durable execution,''
\url{https://temporal.io}. Accessed 2026.
\bibitem{cadence} Uber, ``Cadence: a distributed, scalable, durable workflow orchestrator,''
\url{https://cadenceworkflow.io}. Accessed 2026.
\bibitem{durablefunctions} Microsoft, ``Durable Functions overview,''
\url{https://learn.microsoft.com/azure/azure-functions/durable/}. Accessed 2026.
\end{thebibliography}
\end{document}